\PassOptionsToPackage{table,dvipsnames}{xcolor}
\documentclass{defaltmlr}

\usepackage{amsmath,amsfonts,bm}

\def\secref#1{section~\ref{#1}}

\def\eqref#1{equation~\ref{#1}}

\def\1{\bm{1}}

\DeclareMathAlphabet{\mathsfit}{\encodingdefault}{\sfdefault}{m}{sl}
\SetMathAlphabet{\mathsfit}{bold}{\encodingdefault}{\sfdefault}{bx}{n}

\usepackage{url}
\usepackage[table]{xcolor}
\usepackage[dvipsnames]{xcolor}
\usepackage{graphicx}
\usepackage{geometry}
\usepackage{booktabs}
\usepackage[T1]{fontenc}
\usepackage{fontawesome}
\usepackage{fancyhdr}
\usepackage{xspace}
\usepackage{enumitem}
\usepackage{etoc}
\usepackage{titletoc}
\usepackage{tcolorbox}
\usepackage{listings}

\usepackage{color}
\usepackage[dvipsnames]{xcolor}
\definecolor{JalapenoRed}{RGB}{183,21,64}
\definecolor{Belize}{RGB}{41,128,185}
\definecolor{Amour}{RGB}{238,82,83}
\usepackage{colortbl}

\usepackage{hyperref}

\usepackage[most]{tcolorbox}
\usepackage{enumitem}

\tcbset{
  promptbox/.style={
    title={#1},
    width=\columnwidth,
    breakable,
    enhanced,
    colback=gray!3,
    colframe=black!30,
    coltitle=black,
    fonttitle=\bfseries\footnotesize,
    boxrule=0.8pt,
    arc=1mm,
    boxsep=2pt,
    left=4pt,
    right=4pt,
    top=4pt,
    bottom=4pt,
    before skip=4pt,
    after skip=6pt
  }
}

\newcolumntype{L}[1]{>{\raggedright\arraybackslash}p{#1}}

\usepackage{titlesec}
\usepackage{titletoc}
\usepackage[most]{tcolorbox}
\usepackage{enumitem}

\usepackage[table]{xcolor}

\definecolor{GroundBG}{HTML}{6e8eb7}
\definecolor{ProfileBG}{HTML}{d79a77}
\definecolor{InteractBG}{HTML}{d77434}
\definecolor{OursBG}{HTML}{3e9b4e}

\usepackage[T1]{fontenc}
\usepackage[utf8]{inputenc}
\usepackage{microtype}
\usepackage{inconsolata}
\usepackage{graphicx}

\usepackage[misc]{ifsym}
\usepackage{wrapfig}

\hypersetup{linkcolor=cyan}
\usepackage{mathtools}
\usepackage[normalem]{ulem}
\usepackage{pifont}
\usepackage{bbding}

\usepackage{tcolorbox}
\usepackage{makecell}

\usepackage{tabularx}
\newcommand{\methodname}{ReVision\xspace}
\newcommand{\tablestyle}[2]{%
    \setlength{\tabcolsep}{#1}%
    \renewcommand{\arraystretch}{#2}%
}
\definecolor{nonzero}{HTML}{9B9B9B}    
\definecolor{bestcell}{HTML}{F2D6C9}   
\definecolor{secondcell}{HTML}{FAECE5} 
\definecolor{lightblue}{HTML}{89d8f7}

\makeatletter
\let\orig@fnsymbol\@fnsymbol
\def\@fnsymbol#1{\ifcase#1\or\relax\else\orig@fnsymbol{#1}\fi}
\makeatother

\title{ReVision: Reprogramming Image Generators for Dense Visual Perception}

\author{
\parbox{\textwidth}{
Yitong Yang\textsuperscript{1,2}, \hspace{0.5mm}
Yinglin Wang\textsuperscript{2,\ding{41}}, \hspace{0.5mm}
Xuexin Liu\textsuperscript{2}, \hspace{0.5mm}
Wei Chen\textsuperscript{1}, \hspace{0.5mm}
Zhaoyang Xu\textsuperscript{1}, \hspace{0.5mm}\\
Jiaxin Lin\textsuperscript{1}, \hspace{0.5mm}
Rui Chen\textsuperscript{1}, \hspace{0.5mm}
Shuting He\textsuperscript{2,\ding{41}}, \hspace{0.5mm}
Kai Wang\textsuperscript{1,\ding{41}}
}
}

\affiliation{\textsuperscript{1}Tencent Hy, \textsuperscript{2}Shanghai University of Finance and Economics}

\abstract{
Recent image generators exhibit rich visual priors that can support dense visual perception. However, existing approaches typically encode multiple perception tasks into a shared parameterization, making joint training sensitive to the mixture and weighting of heterogeneous data and prone to cross-task interference. To address this issue, we introduce ReVision, an instruction-conditioned parameter generation framework built on a frozen multimodal understanding model and a frozen image generator. Given an image and a free-form instruction, the understanding model jointly encodes them into a multimodal image--instruction representation, which a trainable parameter generator maps to transient low-rank updates for the generator. Each image--instruction pair induces a distinct set of transient prediction parameters, enabling instance-specific adaptation while both pretrained backbones remain unchanged. ReVision requires no task-specific heads, explicit task identifiers, or stored adapters. A shared RGB output interface unifies segmentation, depth estimation, and surface-normal prediction. Across diverse dense prediction benchmarks, ReVision achieves competitive performance with state-of-the-art methods while supporting flexible task specification through natural-language instructions. Our analysis further reveals substantial gradient disagreement across tasks, highlighting the conflict introduced when heterogeneous objectives share a single parameterization. Meanwhile, the generated updates exhibit structured task- and input-dependent organization in weight space, further validating instance-level parameter generation for multi-task perception.
}

\correspondence{\ding{41} Yinglin Wang (\email{wang.yinglin@shufe.edu.cn}), Shuting He (\email{shuting.he@sufe.edu.cn}), Kai Wang (\email{e0823043@u.nus.edu})}
\website{\url{https://yangyt46.github.io/Revision/}}
\date{\today}

\begin{document}

\maketitle

\makeatletter
\let\@fnsymbol\orig@fnsymbol
\makeatother

\section{Introduction}
\label{sec:intro}

Visual perception requires models to recognize objects, understand scene
semantics, and reason about spatial structure and geometry. Visual
representation learning has traditionally relied on discriminative
supervision~\citep{krizhevsky2012imagenet,dosovitskiy2020image,dehghani2023scaling},
contrastive learning~\citep{chen2020simple,he2020momentum,radford2021learning},
self-distillation~\citep{zhou2022ibot,caron2021emerging}, or
reconstruction-based pretraining~\citep{bao2021beit,he2022masked}. In parallel,
recent advances in image generation~\citep{blackforestlabs2025kontext,
liu2026ernieimage,gao2025seedream,zhao2026qwen}
have substantially expanded what generative models can represent.
SOTA generators synthesize complex scenes with high visual
fidelity, semantic alignment, and fine-grained control, suggesting that their
representations capture objects, geometry, semantics, and their interactions. These works have also demonstrated potential for visual perception.
\begin{figure*}[t]
    \centering
    \includegraphics[width=\linewidth]{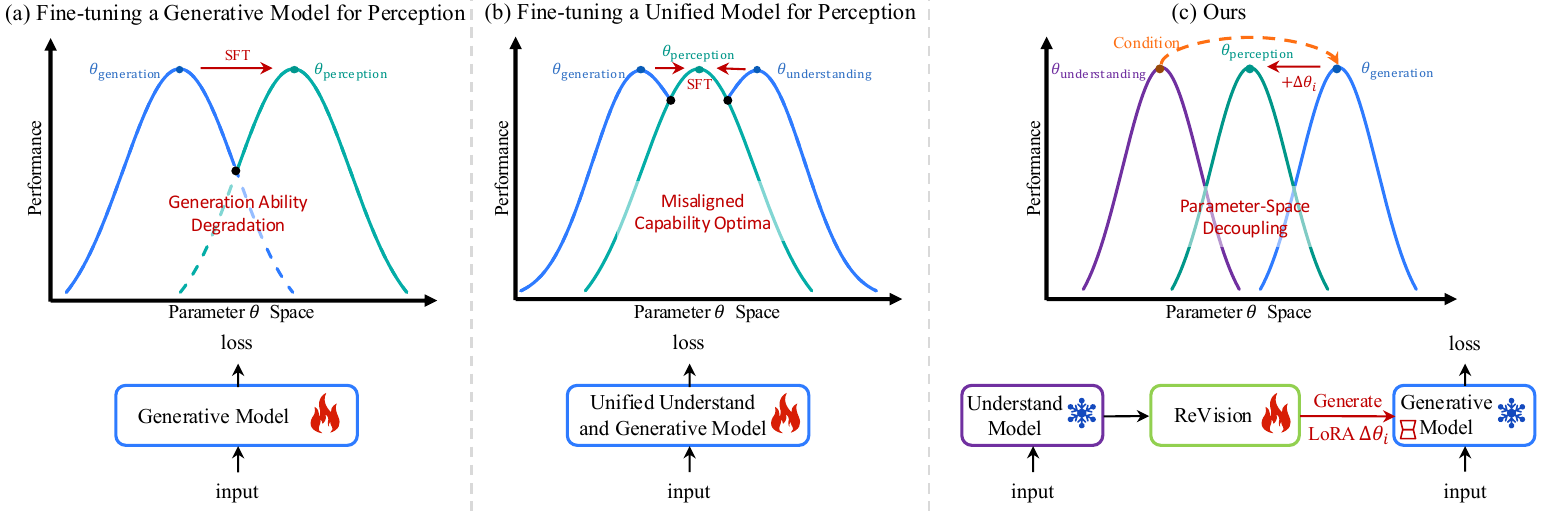}
    \vspace{-15pt}
\caption{\textbf{Parameter-space decoupling for visual perception.}
(a) Fine-tuning repurposes a pretrained generator through a shared set of optimized parameters.
(b) Unified models encode understanding, generation, and perception within one parameterization.
(c) \methodname keeps both backbones frozen and generates instruction-conditioned low-rank updates for each input.}
    \label{fig:f1}
    \vspace{-5pt}
\end{figure*}

Recent work has explored this possibility by repurposing pretrained image
generators for visual perception. Early studies extracted intermediate
diffusion features for segmentation, correspondence, and related recognition
tasks~\citep{zhao2023unleashing,xu2023open,tang2023emergent}. Later methods
formulated dense prediction directly as image generation, adapting pretrained
generators to produce depth maps, surface normals, segmentation masks, and
other spatially structured outputs
~\citep{ke2024repurposing,fu2024geowizard,zhao2026diception,xu2025jodi}.
More recent approaches extend this formulation to multiple dense prediction
tasks through a shared visual output space and natural-language instructions
~\citep{gabeur2026image,zhao2026unipercept}. Related multimodal foundation models combine visual understanding and generation
within a common architecture
~\citep{sun2024emu,wang2024emu3,team2024chameleon,wu2025janus,deng2025bagel},
with SenseNova-Vision further extending this paradigm to heterogeneous vision
tasks using language for symbolic outputs and images for spatial predictions
~\citep{han2026sensenova}. Although these works establish image generation as a viable interface for visual
perception, perception and generation remain coupled through shared model
parameters, as illustrated in Fig.~\ref{fig:f1}. Fine-tuning a pretrained
generator for perception alters the parameter configuration that supports its
original generative capability, often requiring careful balancing between
generation and perception data. Unified understanding--generation models face
a similar limitation, since understanding, generation, and perception must
coexist within a single parameterization even when their preferred solutions
differ. Extending such models therefore still relies on further parameter
optimization. We ask: \emph{Can a pretrained image generator be dynamically
repurposed for visual perception without updating its parameters?}

We answer this question with \methodname, which connects a frozen multimodal
understanding model to a frozen image generator through a trainable parameter
generator. Given an image and a natural-language instruction, the understanding
model produces a multimodal image--instruction representation, from which the
parameter generator synthesizes transient, instance-specific low-rank updates
for the generator. A separate update is generated for each image--instruction
pair and used only for the current prediction, leaving both pretrained
backbones unchanged. Different inputs can therefore instantiate different
predictors without explicit task identifiers, task-specific heads, or stored
adapters. Unlike prior hypernetwork-based methods that condition parameter
generation on image features, predefined tasks, classifier descriptions, or
reusable concepts~\citep{tian2020conditional,nirkin2021hyperseg,zhang2021dodnet,
liu2022polyhistor,tian2025text2weight,li2026sglora,charakorn2025texttolora,
liang2025dragdrop}, \methodname generates parameters from multimodal
image--instruction representations and applies them to a frozen image generator.
We evaluate \methodname on segmentation, depth estimation, and surface-normal
prediction, all expressed in a shared RGB output space, achieving competitive
performance with state-of-the-art methods while supporting flexible task and
output specification through natural-language instructions. We further analyze the generated updates and observe substantial cross-task gradient disagreement under shared-parameter multi-task learning, together with
structured task- and input-dependent organization in weight space, supporting
the effectiveness of instance-level parameter generation. Our main contributions are:
\begin{itemize}[leftmargin=1.4em,itemsep=1.5pt,topsep=2pt]
\item We introduce \methodname, an instruction-conditioned parameter generation
framework that maps multimodal image--instruction representations from a frozen
understanding model to transient, instance-specific low-rank updates for a
frozen image generator.
\item \methodname enables segmentation, depth estimation, and surface-normal
prediction within a shared RGB output space while achieving performance
competitive with state-of-the-art methods without task-specific heads, explicit task identifiers, or stored adapters.
\item Our analysis reveals cross-task gradient conflict under shared-parameter
multi-task learning and structured task- and input-dependent organization of the
generated updates in weight space, supporting instance-level parameter generation.
\end{itemize}

\section{Related Work}
\label{sec:background}
\noindent \textbf{Image Generative Models for Dense Perception.}
Pretrained image generators encode useful priors for geometry, semantics, and spatial structure. Early work exploited intermediate diffusion features for segmentation and correspondence~\citep{zhao2023unleashing,xu2023open,tang2023emergent}, while subsequent methods repurposed generative models for dense prediction, especially depth and surface normals~\citep{ke2024repurposing,fu2024geowizard,xu2025matters,he2025lotus,he2025lotus2}. Recent methods further unify heterogeneous perception tasks through image-space outputs and prompt-based task specification~\citep{zhao2026diception,xu2025jodi,khangaonkar2025gen2seg,zhao2026unipercept}, with stronger generators showing effective perception under limited adaptation~\citep{gabeur2026image,liu2026zeroshot,yang2026rino}. These methods still rely on a fixed set of adapted parameters. ReVision instead keeps the generator frozen and generates an instance-specific parameter update from each image--instruction pair.

\noindent \textbf{Unified Models for Understanding and Generation.}
Prior work has unified visual understanding and generation within a single model. Early approaches cast heterogeneous vision tasks into common sequence or image prediction interfaces~\citep{chen2022pix2seq,lu2023unifiedio,lu2024unifiedio2,mizrahi20234m,bachmann20244m21,wang2023seggpt,geng2024instructdiffusion}. Recent multimodal models integrate understanding and generation in shared architectures~\citep{sun2024emu,wang2024emu3,team2024chameleon}, using shared representations, mixed objectives, or specialized experts to handle their different requirements~\citep{wu2025janus,zhou2025transfusion,xie2025showo,deng2025bagel}. They can also be adapted to heterogeneous perception tasks through instruction tuning~\citep{han2026sensenova}. ReVision keeps both models frozen and uses a parameter generator to translate multimodal representations from the understanding model into instance-specific low-rank updates for the image generator, coupling them through generated parameters rather than shared weights.

\noindent \textbf{Conditional Parameter Generation.}
Hypernetworks generate target-network parameters from conditioning signals, offering an alternative to fixed parameterization~\citep{ha2017hypernetworks,jia2016dynamic}. They have been used for dynamic prediction~\citep{tian2020conditional,wang2020solov2,zhang2021knet}, task-conditioned adaptation~\citep{zhang2021dodnet,liu2022polyhistor}, and lightweight adapter generation~\citep{hu2022lora,mahabadi2021parameter,phang2023hypertuning}. Recent methods predict LoRA weights from language or visual conditions for transfer, classification, and generative personalization~\citep{charakorn2025texttolora,duan2026compressing,tian2025text2weight,li2026sglora,ruiz2024hyperdreambooth,wu2026hy-wu}. ReVision instead generates low-rank updates from free-form image--instruction representations and applies them to a frozen image generator for dense prediction. These updates define an instance-specific perception function rather than a fixed task adapter, classifier head, or personalization module.

\begin{figure*}[t!]
    \centering
    \includegraphics[width=\linewidth]{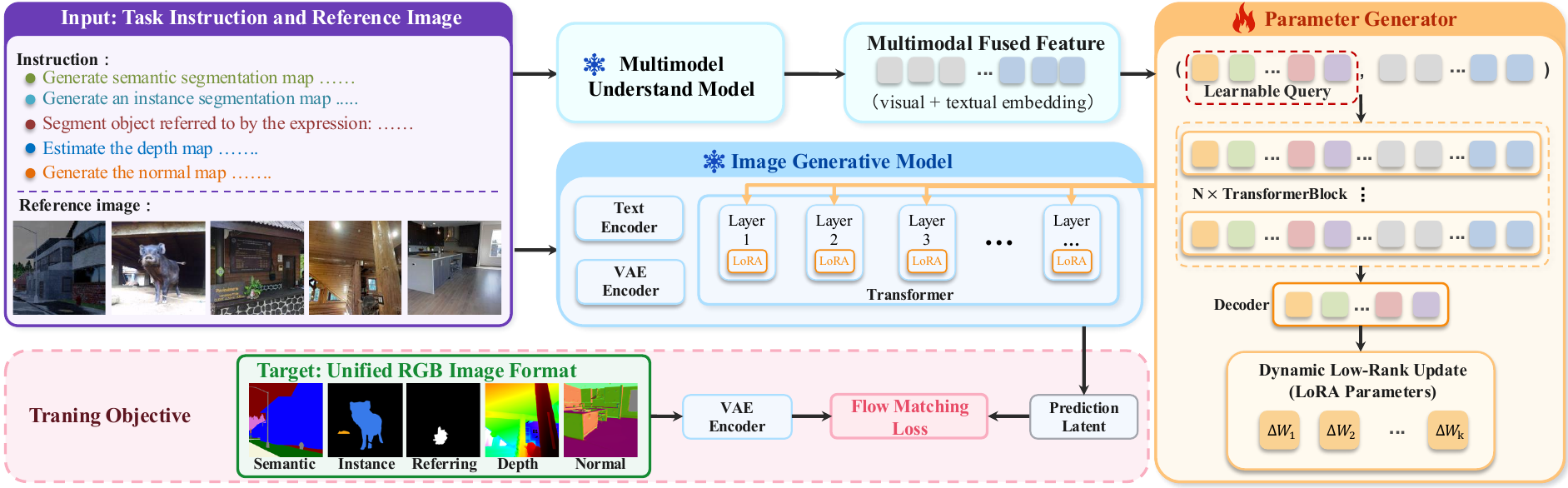}
    \vspace{-10pt}
    \caption{\textbf{Overview of \methodname.} A frozen multimodal model extracts an image--instruction representation that conditions a parameter generator to produce input-dependent low-rank updates for a frozen image generator. Dense perception tasks are unified as RGB generation, with only the parameter generator trained using a flow-matching objective.}
    \label{fig:model}
    \vspace{-15pt}
\end{figure*}
\section{Method}
\label{sec:method}

\subsection{Motivation and Formulation}
\label{sec:method:formulation}

Let $G_\theta$ denote a pretrained image generator with parameters $\theta$, and let a dense-prediction sample be $(I,P,y)$, where $I$ is the input image, $P$ is a free-form natural-language instruction, and $y$ is the target output. Existing approaches typically learn a shared parameter set:
\begin{equation}
\theta^{\star}
=
\arg\min_{\theta}
\mathbb{E}_{(I,P,y)\sim\mathcal{D}}
\left[
\mathcal{L}\big(G_{\theta}(I,P),y\big)
\right].
\label{eq:finetune}
\end{equation}
Although the desired behavior may be specified through an instruction or task token, all inputs are ultimately processed by the same optimized parameters $\theta^\star$. Heterogeneous generation and perception behaviors must therefore coexist within a single parameter configuration, which can introduce interference across tasks. We instead learn a parameter generator $H_{\psi}$ that maps an input-dependent multimodal representation $\mathbf{z}$ to parameter updates for the generator:
\begin{equation}
\theta(\mathbf{z})
=
\theta \oplus H_{\psi}(\mathbf{z}),
\qquad
\hat{y}
=
G_{\theta(\mathbf{z})}(I,P),
\label{eq:ours}
\end{equation}
where $\oplus$ denotes applying the generated update to the pretrained parameters. The parameters $\theta$ remain frozen, while each conditioning representation $\mathbf{z}$ induces a transient parameterization. Different conditions can thus instantiate different predictors without modifying the pretrained generator or storing task-specific parameters. Generating the full parameter set of an image generator is prohibitively expensive, so we generate only LoRA updates~\citep{hu2022lora} for a selected set of layers.

\subsection{Parameter Generator}
\label{sec:method:pg}
\noindent \textbf{Conditioning input.} The parameter generator requires a conditioning representation that captures
both the visual content of the input and the task specified by the instruction.
Although pretrained image generators encode rich visual structure, their
representations are optimized primarily for image synthesis. We therefore use a
separate frozen multimodal understanding model, \texttt{Qwen3.5-9B-Base}~\citep{qwen3.5}, to
provide task-aware representations for perception. These representations
condition the parameter generator, which steers the pretrained visual
representations of the generator toward the requested dense prediction behavior. Specifically, the input image and instruction are jointly processed by the
multimodal understanding model. We extract hidden states from an intermediate
layer, where visual and textual information has been fused while remaining less
specialized toward next-token prediction. The conditioning feature $\mathbf{Z}$ is defined as
\begin{equation}
\mathbf{Z}
=
W_{\mathrm{proj}}\Phi_k(I,P),
\end{equation}
where $\Phi_k(I,P)$ denotes the multimodal hidden states from layer $k$, and
$W_{\mathrm{proj}}$ maps them to the hidden dimension of the parameter
generator. We set $k=16$ in all experiments.

\noindent \textbf{Query-based parameter generation.} To generate LoRA parameters for the image generator, we introduce a bank of learnable parameter queries. Let $S$ denote the number of LoRA injection sites in the frozen image generator, $r$ the LoRA rank, and $d$ the hidden dimension of the parameter generator. We define $Q \in \mathbb{R}^{2Sr \times d},$ where each injection site is assigned $r$ queries for LoRA-$A$ and another $r$
queries for LoRA-$B$. The queries are learned jointly with the parameter generator and serve as site- and factor-specific parameter slots. They are then concatenated with the conditioning feature $\mathbf{Z}$ and processed by $N$ Transformer blocks $\mathcal{T}$:
\begin{equation}
[\mathbf{Q}'; \mathbf{Z}']
=
\mathcal{T}\!\left([\mathbf{Q};\mathbf{Z}]\right).
\end{equation}
Each block consists of parallel multi-head self-attention and gated MLP branches, allowing the parameter queries to interact extensively with the conditioning tokens throughout the Transformer stack. After the final block, we retain only the updated query features $Q' \in \mathbb{R}^{2Sr \times d}$ and discard the transformed conditioning features $Z'$. We then partition $Q'$ according to the injection sites and LoRA factors. For each injection site $s$, the corresponding query features are denoted by
$X_s^A, X_s^B \in \mathbb{R}^{r \times d}$,
which are used to generate the LoRA-$A$ and LoRA-$B$ parameters, respectively.
The two groups are decoded independently by site-specific linear projections,
\begin{equation}
A_s = f_s^A(X_s^A),
\qquad
B_s = f_s^B(X_s^B),
\qquad s=1,\ldots,S,
\end{equation}
where $f_s^A$ and $f_s^B$ map the query features to the dimensions required by
the LoRA-$A$ and LoRA-$B$ matrices at injection site $s$. This results in
$2S$ decoding heads, each producing one LoRA factor for its site.
The generated LoRA parameters are then injected into the frozen image generator.

\noindent \textbf{LoRA injection sites.}
We use \texttt{FLUX.2-klein-base-9B}~\citep{flux2024} as the frozen image
generator, consisting of 8 double-stream and 24 single-stream blocks.
Generated LoRA parameters are injected into selected projection layers
throughout the backbone, with eight sites per double-stream block and two per
single-stream block, yielding $S=112$ injection sites in total.
For each site, the parameter generator produces the corresponding LoRA-$A$ and
LoRA-$B$ matrices. We use rank $r=16$ and set $\alpha/r=1$ for all sites,
while the base generator remains frozen. Detailed model and injection settings are provided in the
Appendix~\ref{app:impl2}.

\subsection{Unified RGB Output Interface}
\label{sec:method:codec}

To keep the frozen generator within its native output space, we follow
\citep{gabeur2026image} and represent all dense-prediction targets as RGB
images. The generator therefore always predicts an RGB image, while the
instruction specifies the desired perception task and output convention.

\noindent \textbf{Segmentation.}
We encode semantic, instance, and referring segmentation as color masks.
For semantic segmentation, each class present in an image is assigned a distinct
color, with the class-to-color mapping explicitly specified in the instruction.
Both the color assignment and its textual representation, including color names,
RGB values, and hexadecimal codes, are randomized across samples to avoid a
fixed dataset-level mapping. For instance segmentation, colors identify
individual object instances and a separate palette is constructed for each
image. For referring segmentation, the referred region and background are
encoded with two contrasting colors specified by the instruction. We retain
only expressions referring to a single instance to ensure an unambiguous target.

\noindent \textbf{Depth estimation.}
We encode relative depth as an RGB image using a continuous color map
$\Gamma:[0,1]\rightarrow[0,1]^3$ defined along a Hamiltonian path through the
eight vertices of the RGB cube. For each image, let $\ell$ and $h$ denote the
$2$nd and $98$th percentiles over valid pixels. We compute
\begin{equation}
u=\operatorname{clip}\!\left(\frac{x-\ell}{h-\ell},0,1\right),\quad
\rho=\tau+(1-\tau)u,\quad
\Gamma(\rho)=v_j+(7\rho-j)(v_{j+1}-v_j),\quad
j=\lfloor 7\rho\rfloor,
\end{equation}
where $x$ denotes depth or inverse depth, oriented such that $u$ increases with
distance, and $\tau>0$ reserves black for invalid pixels. The predicted RGB
value is decoded by projection onto the same path. This per-image normalization
removes absolute scale and shift, allowing metric and scale-ambiguous sources
to be trained jointly. At evaluation time, predictions are aligned to the
ground truth by least-squares scale and shift. We discard samples with
negligible relative depth variation,
$(h-\ell)/\max(|h|,|\ell|)<\epsilon$, to avoid amplifying sensor or
reconstruction noise.

\noindent \textbf{Surface normal estimation.}
Surface normals are encoded directly as RGB images by linearly mapping each
component from $[-1,1]$ to $[0,255]$. Predictions are mapped back to
$[-1,1]$ and normalized to unit length, while invalid pixels are encoded as
black. Normals from all sources are converted to a common camera coordinate
system, with $X_{\mathrm{cam}}$ pointing right, $Y_{\mathrm{cam}}$ down, and
$Z_{\mathrm{cam}}$ into the scene. Since surface normals are scale invariant,
both metric and scale-ambiguous geometry sources can be used jointly for
training. Data generation details are provided in the Appendix~\ref{app:data}.

\subsection{Training}
\label{sec:method:training}

\noindent \textbf{Objective.} Given noise $\varepsilon \sim \mathcal{N}(0,I)$ and timestep $t \in (0,1)$, let $\mathbf{e}^{\mathrm{gt}}$ denote the VAE-encoded ground-truth output, $\mathbf{e}_{\mathrm{ref}}$ the VAE-encoded reference image $I$, and $\mathbf{e}_{p}$ the text embedding of the instruction $P$. We further denote by $x_t$ the denoising target at timestep $t$. With the image generator frozen, we only optimize the parameter generator $H_\psi$:
\begin{equation}
\mathcal{L}(\psi)
=
\Big\lVert\
G_{\theta\oplus H_\psi(\mathbf{Z})}
\left(
\mathbf{x}_t,
\mathbf{e}_{\mathrm{ref}},
\mathbf{e}_{p},
t
\right)
-
(\varepsilon - \mathbf{e}^{gt})
\Big\rVert_2^2 .
\label{eq:loss}
\end{equation}
We sample $t$ from a logit-normal distribution~\citep{esser2024scaling} and apply a fixed resolution-independent shift of $e^{0.8}\approx2.2255$.

\noindent \textbf{Training schedule.}
We train progressively by resolution, using $256^2$ for $10$k steps on four tasks, followed by $512^2$ for $20$k steps with referring segmentation added. We transfer the raw hypernetwork weights between stages and restart the optimizer and learning-rate schedule. Both stages use AdamW with a peak learning rate of $2\times10^{-5}$, uniform task sampling, and mixed-task batches. We maintain an EMA with decay $0.999$ and use the EMA checkpoint for evaluation. Full hyperparameters are provided in Appendix~\ref{app:impl}.

\noindent \textbf{Training data.} All tasks are drawn from public corpora and tiled into fixed-resolution crops. Semantic segmentation data are collected from ADE20K~\citep{zhou2017ade20k}, COCO-Stuff-164K~\citep{caesar2018cocostuff}, Mapillary Vistas v2~\citep{neuhold2017mapillary}, and BDD7K~\citep{yu2020bdd100k}, while instance segmentation uses LVIS~\citep{gupta2019lvis}, Open Images V7~\citep{kuznetsova2020openimages}, and GranD~\citep{rasheed2024glamm}. We relabel these datasets with SAM 3~\citep{carion2026sam}, followed by deduplication and quality filtering, to obtain fine-grained masks. Referring segmentation uses the original instructions from GranD~\citep{rasheed2024glamm} together with SAM 3 relabeling. Depth estimation and surface-normal estimation are treated as separate tasks, both using Omnidata~\citep{eftekhar2021omnidata} and TartanAir~\citep{wang2020tartanair}. No evaluation dataset is included in training.

\section{Experiments}
\label{sec:exp}

\subsection{Comparison Results}
For a controlled comparison, we additionally train two variants using the same
image generator. \textit{Task-specific LoRA} trains one adapter per task, while
\textit{Multi-task LoRA} jointly trains a single adapter on all tasks.
\methodname uses the same multi-task data but generates task- and
input-dependent updates from the instruction. Qualitative results are provided in Appendix~\ref{app:q_result}.

\begin{wraptable}[11]{r}{0.45\columnwidth}
\vspace{-15pt}
\caption{Semantic segmentation on Cityscapes.}
\vspace{-15pt}
\vspace{2mm}
\label{tab:cityscapes}
\centering
\tablestyle{4pt}{1.05}
\resizebox{1.0\linewidth}{!}{%
\begin{tabular}{lcc}
\toprule
Method & Zero-shot & mIoU $\uparrow$ \\
\midrule

\textcolor{nonzero}{SegMan-L~\citep{fu2025segman}}
    & \textcolor{nonzero}{\ding{55}}
    & \textcolor{nonzero}{84.2} \\
\textcolor{nonzero}{SenseNova-Vision~\citep{han2026sensenova}}
    & \textcolor{nonzero}{\ding{55}}
    & \textcolor{nonzero}{71.2} \\

\midrule

APE-D~\citep{shen2024aligning}
    & \ding{51} & 44.2 \\
OpenSeeD~\citep{zhang2023simple}
    & \ding{51} & 47.8 \\
X-Decoder~\citep{zou2023generalized}
    & \ding{51} & 52.0 \\
SAM 3~\citep{carion2026sam}
    & \ding{51} & 65.2 \\
Vision-Banana~\citep{gabeur2026image}
    & \ding{51} & 69.9 \\

\midrule

\textit{Task-specific LoRA}
    & \ding{51} & \underline{70.07} \\
\textit{Multi-task LoRA}
    & \ding{51} & 64.68 \\

\rowcolor{cyan!8} \textbf{\methodname} (Ours)
    & \ding{51} & \textbf{72.75} \\

\bottomrule
\end{tabular}%
}
\end{wraptable}
\noindent \textbf{Semantic Segmentation.}
We evaluate zero-shot semantic segmentation on Cityscapes val., whose images,
19-class \texttt{trainId} taxonomy, and label conventions are absent from
training. At inference time, both the target categories and their color
mappings are specified through natural-language instructions, allowing
open-vocabulary task specification without a fixed label set. As shown in
Tab.~\ref{tab:cityscapes}, \methodname reaches $72.75$ mIoU and outperforms
both LoRA baselines. The lower performance of Multi-task LoRA relative to
Task-specific LoRA also suggests interference when heterogeneous tasks share a
single fixed update.

\begin{wraptable}[14]{r}{0.48\columnwidth}
\vspace{-5pt}
\centering
\caption{Instance segmentation on SA-Co/Gold.}
\vspace{-10pt}
\label{tab:zero-shot-transfer}
\vspace{2mm}
\tablestyle{2.5pt}{1.05}

\resizebox{1.0\linewidth}{!}{%
\begin{tabular}{lcccc}
\toprule
Model & Zero-shot & $cgF_1 \uparrow$ & IL\_MCC $\uparrow$ & $pmF_1 \uparrow$ \\
\midrule

\textcolor{nonzero}{SAM 3~\citep{carion2026sam}}
    & \textcolor{nonzero}{\ding{55}}
    & \textcolor{nonzero}{54.1}
    & \textcolor{nonzero}{0.82}
    & \textcolor{nonzero}{66.1} \\

\textcolor{nonzero}{
    \makecell[l]{
        SAM 3~\citep{carion2026sam}\\
        + Llama 3.2 (ft)
    }
}
    & \textcolor{nonzero}{\ding{55}}
    & \textcolor{nonzero}{61.2}
    & \textcolor{nonzero}{0.86}
    & \textcolor{nonzero}{70.8} \\

\midrule

gDino-T~\citep{liu2024grounding}
    & \ding{51} & 3.3 & 0.15 & 16.2 \\

LLMDet-L~\citep{fu2025llmdet}
    & \ding{51} & 6.5 & 0.21 & 27.3 \\

Gemini 2.5~\citep{comanici2025gemini}
    & \ding{51} & 13.0 & 0.29 & 46.1 \\

APE-D~\citep{shen2024aligning}
    & \ding{51} & 16.4 & 0.40 & 36.9 \\

DINO-X~\citep{ren2024dino}
    & \ding{51}
    & 21.3
    & 0.38
    & \underline{55.2} \\

OWLv2~\citep{minderer2023scaling}
    & \ding{51} & 24.6 & 0.57 & 42.0 \\

\makecell[l]{
    Vision Banana~\citep{gabeur2026image}\\
    + Gemini 3.1 Flash-Lite
}
    & \ding{51}
    & \textbf{47.5}
    & \textbf{0.84}
    & \textbf{56.0} \\

\midrule

\textit{Task-specific LoRA} + Qwen3.5-9B
    & \ding{51}
    & 33.95
    & \underline{0.74}
    & 45.7 \\

\textit{Multi-task LoRA} + Qwen3.5-9B
    & \ding{51}
    & 30.71
    & \underline{0.74}
    & 41.4 \\

\rowcolor{cyan!8} \textbf{\methodname} (Ours) + Qwen3.5-9B
    & \ding{51}
    & \underline{40.17}
    & \underline{0.74}
    & 54.3 \\

\bottomrule
\end{tabular}%
}
\end{wraptable}
\noindent \textbf{Instance Segmentation.}
We further evaluate on the open-vocabulary SA-Co/Gold benchmark, where the model must separate an unknown number of instances of the queried concept. As shown in Tab.~\ref{tab:zero-shot-transfer}, \methodname achieves $40.17$ $cgF_1$, $0.74$ IL\_MCC, and $54.3$ $pmF_1$, outperforming both LoRA baselines and several representative zero-shot methods. The gains are largest on $cgF_1$ and $pmF_1$, indicating improved instance separation and mask quality. IL\_MCC is similar across our three variants, suggesting that instance-existence prediction is mainly determined by the shared \texttt{Qwen3.5-9B} component. A gap remains to Vision Banana, particularly on $cgF_1$, likely reflecting differences in the underlying foundation model.

\begin{wraptable}[13]{r}{0.48\columnwidth}
\centering
\caption{Referring expression segmentation results on RefCOCOg val (UMD).}
\label{tab:zero-shot-seg}
\vspace{-10pt}
\vspace{2mm}
\tablestyle{3pt}{1.05}
\resizebox{\linewidth}{!}{%
\begin{tabular}{lcc}
\toprule
Model & Zero-shot & cIoU $\uparrow$ \\
\midrule

\textcolor{nonzero}{HyperSeg-Phi2-2.7B~\citep{wei2024hyperseg}}
    & \textcolor{nonzero}{\ding{55}}
    & \textcolor{nonzero}{79.4} \\
\textcolor{nonzero}{X-SAM-Phi3-3.8B~\citep{wang2026x}}
    & \textcolor{nonzero}{\ding{55}}
    & \textcolor{nonzero}{83.8} \\
\textcolor{nonzero}{SenseNova-Vision~\citep{han2026sensenova}}
    & \textcolor{nonzero}{\ding{55}}
    & \textcolor{nonzero}{80.3} \\

\midrule

HybridGL~\citep{liu2025hybrid}
    & \ding{51} & 51.3 \\
LocalizationHeads-LLaVA-1.5-13B~\citep{kang2025your}
    & \ding{51} & 67.7 \\
SAM 3~\citep{carion2026sam} + Gemini 2.5 Pro
    & \ding{51} & \underline{73.4} \\
Vision Banana~\citep{gabeur2026image}
    & \ding{51} & \textbf{73.8} \\

\midrule

\textcolor{nonzero}{\textit{RefCOCOg-supervised LoRA}}
    & \textcolor{nonzero}{\ding{55}}
    & \textcolor{nonzero}{54.7} \\
\textit{Task-specific LoRA}
    & \ding{51} & 53.1 \\
\textit{Multi-task LoRA}
    & \ding{51} & 46.8 \\
\rowcolor{cyan!8} \textbf{\methodname} (Ours)
    & \ding{51} & 53.8 \\

\bottomrule
\end{tabular}%
}
\end{wraptable}

\noindent \textbf{Referring Segmentation.}
Referring segmentation identifies a target from a free-form description,
requiring finer grounding and relational reasoning than fixed-category segmentation. As shown in
Tab.~\ref{tab:zero-shot-seg}, \methodname reaches $53.8$ cIoU, outperforming
both LoRA baselines and approaching the RefCOCOg-supervised LoRA baseline.
This result highlights the ability of instance-conditioned parameter generation
to translate open-ended language descriptions into pixel-level predictions.
A gap remains to Vision Banana, which may stem from the stronger base generator. In particular,
GenMask~\citep{yang2026genmask} shows that mask generation benefits from
segmentation-specific noise and timestep sampling, whereas \methodname retains
the original generative schedule.

\begin{table*}[h!]
\tablestyle{3.5pt}{0.92}
\centering
\caption{\textbf{Comparison on monocular depth estimation.}
$^\ast$SenseNova-Vision is not strictly zero-shot due to dense geometry supervision and possible evaluation-source overlap. $^\ddagger$Vision Banana reports metric-depth without per-image affine alignment and is included for reference.}
\label{tab:depth}
\vspace{-2mm}
\resizebox{\linewidth}{!}{%
\begin{tabular}{lccccccc}
\toprule
\multirow{2}{*}{Method}
& \multirow{2}{*}{Zero-shot}
& \multicolumn{6}{c}{AbsRel $\downarrow$ / $\delta_1 \uparrow$} \\
\cmidrule(lr){3-8}
& & NYUv2 & KITTI & ETH3D & ScanNet & DIODE & iBims1 \\
\midrule

Marigold~\citep{ke2024repurposing}
& \ding{51}
& 5.5 / 96.4
& 9.9 / 91.6
& 6.5 / 95.9
& 6.4 / 95.2
& 30.8 / 77.3
& - / - \\

DICEPTION~\citep{zhao2026diception}
& \ding{51}
& 6.1 / 96.0
& 6.9 / \underline{94.9}
& 5.0 / 97.5
& 7.2 / 94.4
& 28.9 / 72.2
& - / - \\

FE2E~\citep{wang2025editor}
& \ding{51}
& \underline{4.1} / \textbf{97.7}
& \textbf{6.6} / \textbf{96.0}
& \underline{3.8} / \textbf{98.7}
& 4.4 / 97.5
& \underline{22.8} / \textbf{81.2}
& - / - \\

Lotus-2~\citep{he2025lotus2}
& \ding{51}
& \underline{4.1} / \underline{97.6}
& \underline{6.7} / 94.5
& 4.6 / \underline{98.1}
& \underline{4.2} / \underline{97.6}
& \textbf{22.1} / 75.2
& - / - \\

Vision Banana$^\ddagger$~\citep{gabeur2026image}
& \ding{51}
& 8.1 / 94.8
& 10.7 / 91.5
& 10.3 / 93.5
& - / -
& - / -
& 7.8 / 93.4 \\

\textcolor{nonzero}{SenseNova-Vision~\citep{han2026sensenova}}
& \textcolor{nonzero}{\ding{55}$^\ast$}
& \textcolor{nonzero}{4.0 / 98.1}
& \textcolor{nonzero}{5.9 / 95.9}
& \textcolor{nonzero}{4.3 / 97.4}
& \textcolor{nonzero}{3.9 / 98.0}
& \textcolor{nonzero}{20.6 / 76.4}
& \textcolor{nonzero}{- / -} \\

\midrule

\textit{Task-specific LoRA}
& \ding{51}
& 5.7 / 95.5
& 22.7 / 70.8
& 5.2 / 96.9
& 5.1 / 96.8
& 36.3 / 72.6
& \underline{4.9} / \underline{96.4} \\

\textit{Multi-task LoRA}
& \ding{51}
& 5.9 / 95.3
& 19.4 / 75.4
& 6.5 / 95.3
& 5.0 / 97.0
& 35.3 / 73.2
& 5.5 / 95.8 \\

\rowcolor{cyan!8} \textbf{ReVision}
& \ding{51}
& \textbf{4.0} / 97.5
& 9.1 / 93.2
& \textbf{3.7} / \underline{98.1}
& \textbf{3.9} / \textbf{98.0}
& 24.5 / \underline{79.6}
& \textbf{3.9} / \textbf{98.0} \\

\bottomrule
\end{tabular}%
}
\end{table*}
\noindent \textbf{Depth Estimation.}
Generated depth images are decoded and evaluated with affine-invariant metrics.
As shown in Tab.~\ref{tab:depth}, \methodname generalizes well across indoor
and outdoor benchmarks, with strong results on NYUv2, ETH3D, ScanNet, and
iBims-1. Task-specific LoRA is competitive on several indoor datasets but
degrades on outdoor datasets such as KITTI and DIODE, whereas Multi-task LoRA
shows better cross-domain robustness. \methodname further improves on both,
combining strong in-domain accuracy with better transfer across scene
distributions.

\noindent \textbf{Surface Normal Estimation.}
We evaluate surface normals using mean and median angular errors. As shown in
Tab.~\ref{tab:normal-estimation}, \methodname achieves strong zero-shot
performance across both indoor and outdoor datasets, with particularly low
errors on NYUv2, ScanNet, and VKitti. It improves the mean error over both LoRA
baselines on nearly all datasets and achieves the best mean and median errors
among our variants on VKitti. Compared with Multi-task LoRA, the gains are most
pronounced under domain shift, indicating more robust transfer across scene
distributions.

\begin{table*}[t]
\tablestyle{3.5pt}{0.95}
\centering
\caption{\textbf{Comparison on surface normal estimation.}
We report mean and median angular errors ($\downarrow$).
$^\ast$Lotus-2-Normal is trained on Virtual KITTI 2, so its VKitti result is not zero-shot.
$^\ddagger$SenseNova-Vision includes ScanNetV2 in training and is not strictly benchmark-disjoint.}
\label{tab:normal-estimation}
\vspace{-2mm}
\resizebox{\textwidth}{!}{%
\begin{tabular}{lccccccccccc}
\toprule
\multirow{3}{*}{Methods}
& \multirow{3}{*}{Zero-shot}
& \multicolumn{8}{c}{Indoor}
& \multicolumn{2}{c}{Outdoor} \\
\cmidrule(lr){3-10}
\cmidrule(lr){11-12}

&
& \multicolumn{2}{c}{Average}
& \multicolumn{2}{c}{NYUv2}
& \multicolumn{2}{c}{DIODE-indoor}
& \multicolumn{2}{c}{ScanNet}
& \multicolumn{2}{c}{VKitti} \\

&
& mean $\downarrow$ & median $\downarrow$
& mean $\downarrow$ & median $\downarrow$
& mean $\downarrow$ & median $\downarrow$
& mean $\downarrow$ & median $\downarrow$
& mean $\downarrow$ & median $\downarrow$ \\
\midrule

Marigold~\citep{ke2024repurposing}
& \ding{51}
& 19.606 & 11.828
& 20.864 & 11.134
& 16.671 & 12.084
& 21.284 & 12.268
& -- & -- \\

DSINE~\citep{bae2024dsine}
& \ding{51}
& 17.017 & 10.190
& \underline{16.4}
& \underline{8.4}
& 18.453 & 13.871
& 16.2 & 8.3
& 28.9 & \underline{9.9} \\

StableNormal~\citep{ye2024stablenormal}
& \ding{51}
& 17.168 & 10.028
& 19.707 & 10.527
& \textbf{13.701}
& \textbf{9.46}
& 18.098 & 10.097
& -- & -- \\

\textcolor{nonzero}{Lotus-2-Normal~\citep{he2025lotus2}}
& \textcolor{nonzero}{\ding{55}$^\ast$}
& \textcolor{nonzero}{16.558}
& \textcolor{nonzero}{--}
& \textcolor{nonzero}{16.9}
& \textcolor{nonzero}{N/A}
& \textcolor{nonzero}{18.575}
& \textcolor{nonzero}{N/A}
& \textcolor{nonzero}{14.2}
& \textcolor{nonzero}{N/A}
& \textcolor{nonzero}{28.894}
& \textcolor{nonzero}{9.677} \\

\textcolor{nonzero}{SenseNova-Vision~\citep{han2026sensenova}}
& \textcolor{nonzero}{\ding{55}$^\ddagger$}
& \textcolor{nonzero}{14.167}
& \textcolor{nonzero}{--}
& \textcolor{nonzero}{14.4}
& \textcolor{nonzero}{--}
& \textcolor{nonzero}{15.3}
& \textcolor{nonzero}{--}
& \textcolor{nonzero}{12.8}
& \textcolor{nonzero}{--}
& \textcolor{nonzero}{--}
& \textcolor{nonzero}{--} \\

Vision Banana~\citep{gabeur2026image}
& \ding{51}
& \underline{15.549}
& \underline{9.300}
& 17.778 & 8.876
& \underline{13.818}
& \underline{11.556}
& 15.052 & 7.468
& 29.063 & 10.699 \\

\midrule

\textit{Task-specific LoRA}
& \ding{51}
& 16.781 & 9.341
& 18.39 & 8.578
& 16.997 & 12.09
& 14.957 & 7.355
& \underline{28.213}
& 12.221 \\

\textit{Multi-task LoRA}
& \ding{51}
& 16.698 & 9.379
& 18.11 & 8.675
& 17.385 & 12.403
& \underline{14.6}
& \underline{7.089}
& 30.664 & 15.638 \\

\rowcolor{cyan!8} \textbf{ReVision}
& \ding{51}
& \textbf{15.467}
& \textbf{9.162}
& \textbf{16.202}
& \textbf{7.994}
& 16.711 & 12.85
& \textbf{13.489}
& \textbf{6.644}
& \textbf{24.854}
& \textbf{8.43} \\

\bottomrule
\end{tabular}%
}
\end{table*}

\subsection{Analyses and Discussions}
\label{sec:analysis}

The key premise of \methodname is that perception should be represented by a
conditional family of parameter updates rather than a single shared parameter
point. To examine whether this structure emerges in practice, we analyze the
LoRA updates generated for five training tasks: semantic segmentation, instance
segmentation, referring segmentation, depth estimation, and surface-normal
estimation. We sample 120 inputs per task, yielding 600 updates in total, and
study them from three perspectives: task-level behavior, gradient interactions,
and parameter-space geometry. 

\noindent \textbf{From Static Adaptation to Conditional Updates.} Tabs.~\ref{tab:cityscapes}--\ref{tab:normal-estimation} compare \methodname with Task-specific LoRA and Multi-task LoRA across five perception tasks. On segmentation, Task-specific LoRA generally outperforms Multi-task LoRA, indicating interference from a single shared update across heterogeneous objectives. In contrast, Multi-task LoRA performs better on several geometric benchmarks and transfers more robustly under distribution shift, suggesting that joint training can capture useful cross-task structure. \methodname improves over both baselines across most settings, with clear gains on segmentation and more stable performance on depth and surface normals. These results highlight the limitation of static updates: task-specific LoRA specializes one parameter set per task, while multi-task LoRA shares one across all tasks. \methodname instead generates a distinct parameter update for each image--instruction pair, allowing different samples to occupy different points in weight space while retaining the benefits of joint multi-task training.

\begin{figure*}[t!]
  \centering
  \includegraphics[width=0.97\linewidth]{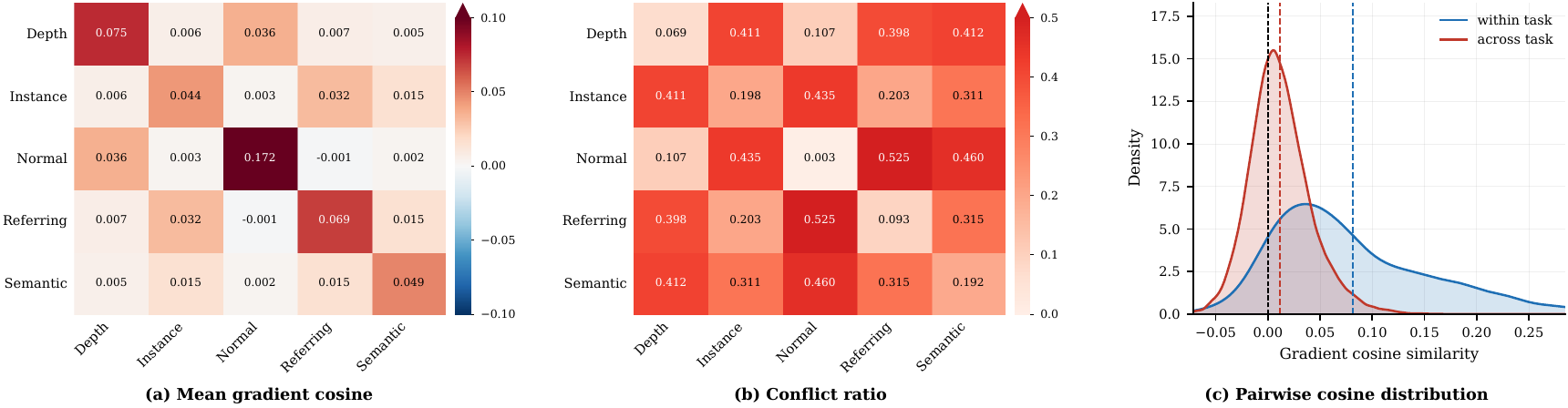}
  \vspace{-10pt}
\caption{\textbf{Gradient geometry across perception tasks.}
\textbf{(a)} Mean pairwise cosine similarity between gradients of the generated LoRA parameters.
\textbf{(b)} Conflict ratio, defined as the fraction of gradient pairs with negative cosine similarity.
\textbf{(c)} Cosine-similarity distributions for within-task and cross-task pairs.
Dashed lines mark zero and the distribution means.}
\vspace{-10pt}
  \label{fig:analysis-gradient}
\end{figure*}
\noindent \textbf{Cross-task Conflict in Gradient Space.} We next ask whether the limitations of static multi-task updates are reflected
in optimization geometry. We fix the flow timestep and noise realization,
treat the ReVision-generated $(A,B)$ as differentiable leaf tensors, and
compute loss gradients with respect to these generated LoRA parameters after
one flow step, without updating the backbone or parameter generator. As shown in Fig.~\ref{fig:analysis-gradient}, the mean cosine similarity drops
from $0.082$ within tasks to $0.012$ across tasks, while the conflict ratio
rises from $11.1\%$ to $35.8\%$. The degree of interference is task dependent:
depth and surface normals remain relatively aligned ($0.036$ cosine,
$10.7\%$ conflict), as do instance and referring segmentation ($0.032$,
$20.3\%$), whereas several cross-family pairs conflict substantially more.
Referring segmentation and surface normals, for example, reach a $52.5\%$
conflict ratio. These results explain why a single multi-task update can become
a compromise: it must reconcile gradients that vary substantially across tasks.
ReVision instead generates a distinct update for each image--instruction
instance, avoiding the need to collapse these directions into one shared
parameter solution.

\begin{table*}[!t]
\tablestyle{2.6pt}{1.06}
\centering
\vspace{-6mm}
\caption{\textbf{Condition-routing ablation.}
Only the parameter-generator condition is perturbed; image-generator inputs remain unchanged. We shuffle text or images within a task to break sample-level correspondence, and route conditions across tasks to break task-level correspondence. SA-Co/Gold, depth, and normals are omitted due to similar text conditions.}
\label{tab:analysis-routing}
\vspace{-2mm}
\resizebox{1.0\linewidth}{!}{%
\begin{tabular}{lccccccccc}
\toprule
Method
& \multicolumn{3}{c}{Cityscapes}
& \multicolumn{1}{c}{SA-Co/Gold}
& \multicolumn{1}{c}{RefCOCOg}
& \multicolumn{2}{c}{Depth}
& \multicolumn{2}{c}{Normal} \\
\cmidrule(lr){2-4}
\cmidrule(lr){5-5}
\cmidrule(lr){6-6}
\cmidrule(lr){7-8}
\cmidrule(lr){9-10}
& mIoU $\uparrow$ & mAcc $\uparrow$ & pixAcc $\uparrow$
& $pmF_1 \uparrow$
& cIoU $\uparrow$
& AbsRel $\downarrow$ & $\delta_1 \uparrow$
& mean$^\circ \downarrow$ & med.$^\circ \downarrow$ \\
\midrule
Base Image Generator
& 20.29 & 36.79 & 59.91
& 4.32
& 7.64
& 0.5522 & 0.4662
& 105.84 & 109.21 \\
\midrule
Shuffled Text (within task)
& 67.74 &79.19 & 94.20
& -
& 46.9 & -
& - & - &-\\
Shuffled Image (within task)
& 68.03 & 79.34 & 94.21
& 48.5
& 51.7
& 0.062 & 0.950
& 18.83 & 11.83 \\
Shuffled condition (cross task)
& 22.16 & 42.65 & 51.71
& 19.6
& 15.7
& 5.85 & 0.286
& 93.61 & 96.00 \\
\midrule
\rowcolor{cyan!8}
ReVision (Ours)
& \textbf{70.27} & \textbf{81.36} & \textbf{94.79}
& \textbf{51.1}
& \textbf{54.2}
& \textbf{0.049} & \textbf{0.971}
& \textbf{17.99} & \textbf{11.02} \\
\bottomrule
\end{tabular}}
\end{table*}
\begin{figure}[t]
\centering
\includegraphics[width=0.7\columnwidth]{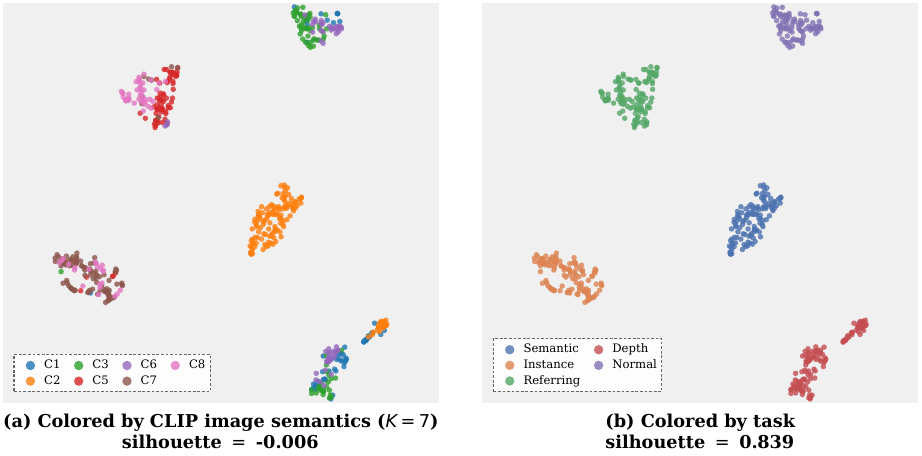}
\caption{\textbf{Generated updates in parameter space.}
The same t-SNE embedding of 7,168-dimensional LoRA update sketches is colored
by (a) CLIP image-semantic clusters ($K=7$) and (b) task identity.
Silhouette scores are shown below each panel.}
\label{fig:analysis-manifold}
\end{figure}
\noindent \textbf{Task Structure with Instance-level Specialization.} 
We examine the organization of generated updates in parameter space. Each LoRA
update is represented by a 7,168-dimensional sketch and visualized with t-SNE.
In Fig.~\ref{fig:analysis-manifold}(a), coloring the embedding by independent
CLIP image-semantic clusters yields little separation
(silhouette $=-0.006$). In contrast, task labels form well-separated regions
(silhouette $=0.839$; Fig.~\ref{fig:analysis-manifold}(b)), indicating that the
coarse geometry is primarily task-structured. Task identity alone, however,
does not determine the update. As shown in Tab.~\ref{tab:analysis-routing},
breaking image--instruction correspondence within a task degrades performance,
while cross-task routing causes substantially larger failures. This suggests a
two-level organization: task determines the coarse region of parameter
space, while the conditioning input selects an instance-specific
update within that region.

\begin{wraptable}{r}{0.5\columnwidth}
\vspace{-8pt}
\centering
\caption{\textbf{Geometry of the generated update family.}
\textbf{Centroid norm} measures the magnitude of the mean update, and \textbf{mean distance} measures the spread around it. Their ratio defines \textbf{dispersion}, the normalized update spread. \textbf{Offset/spread} measures task separation from the global centroid relative to within-task variation, while \textbf{participation ratio} estimates the effective dimensionality. We also report the effective rank and relative Frobenius norm $\lVert\Delta W_s\rVert_F/\lVert W_s\rVert_F$ of the generated LoRA updates. Detailed definitions are provided in the Appendix~\ref{app:analysis}.}
\label{tab:analysis-dispersion}
\vspace{2mm}
\setlength{\tabcolsep}{2.0pt}
\renewcommand{\arraystretch}{0.94}
\scriptsize

\resizebox{\linewidth}{!}{%
\begin{tabular}{lcccccc}
\toprule
& Sem. & Inst. & Ref. & Depth & Normal & All \\
\midrule
Centroid norm
& 359.9 & 380.2 & 392.8 & 352.7 & 330.2 & 316.9 \\
Mean distance
& 8.6 & 25.1 & 13.4 & 75.2 & 92.6 & 181.6 \\
Dispersion
& 0.024 & 0.066 & 0.034 & 0.213 & 0.281 & 0.573 \\
Offset / spread
& 14.8 & 5.5 & 11.9 & 2.8 & 2.5 & -- \\
Participation ratio
& 2.12 & 3.08 & 5.85 & 2.48 & 1.84 & 1.91 \\
\midrule
Effective rank (mean)
& 13.20 & 13.20 & 13.14 & 13.61 & 13.65 & 13.36 \\
Effective rank (min)
& 3.72 & 3.99 & 4.04 & 3.68 & 3.52 & 3.52 \\
Effective rank (max)
& 15.94 & 15.95 & 15.95 & 15.95 & 15.95 & 15.95 \\
\midrule
$\lVert\Delta W_s\rVert_F/\lVert W_s\rVert_F$ (mean) & 18.54 & 19.53 & 19.97 & 18.04 & 17.19 & 18.65 \\
$\lVert\Delta W_s\rVert_F/\lVert W_s\rVert_F$ (min)
&  1.64 &  1.73 &  1.80 &  1.34 &  1.24 &  1.24 \\
$\lVert\Delta W_s\rVert_F/\lVert W_s\rVert_F$ (max)
& 37.39 & 40.22 & 39.56 & 44.24 & 47.32 & 47.32 \\
\bottomrule
\end{tabular}%
}
\vspace{-8pt}
\end{wraptable}
\noindent \textbf{Geometry and Backbone Allocation of Generated Updates:}
\textbf{(1) Task-structured update geometry.}
Tab.~\ref{tab:analysis-dispersion} shows clear task-dependent geometry in the generated updates. Within-task dispersion ($0.024$--$0.281$) is much smaller than that across all samples ($0.573$), while task centroids lie $2.5$--$14.8$ within-task spreads from the global centroid, indicating distinct task-specific regions. Segmentation updates are particularly compact, whereas depth and surface normals exhibit greater variation. The effective rank and relative update magnitude remain stable across tasks, suggesting that task specialization is expressed primarily through the geometry of the generated updates rather than changes in their rank or scale.
\begin{figure*}[t]
  \centering
  \vspace{-10pt}
  \includegraphics[width=0.98\linewidth]{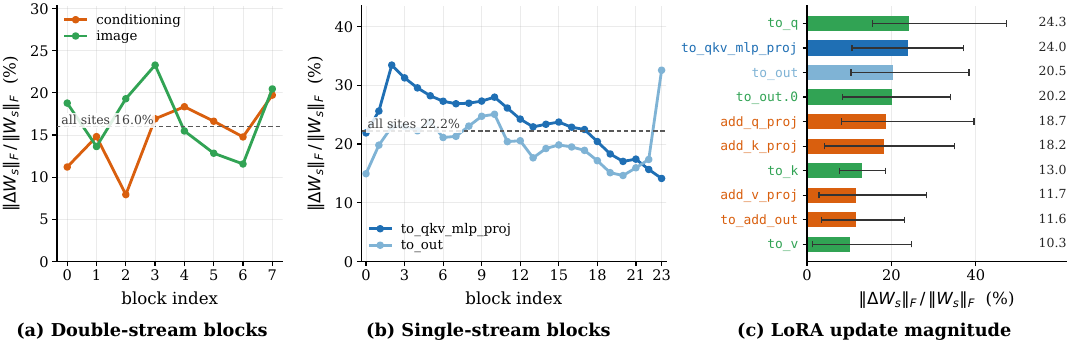}
\caption{\textbf{Layerwise and module-wise profile of the generated LoRA updates.}
Relative update magnitude $\lVert\Delta W_s\rVert_F/\lVert W_s\rVert_F$
across \textbf{(a)} double-stream and \textbf{(b)} single-stream blocks, with
dashed lines indicating the all-site mean. \textbf{(c)} Mean relative update
magnitude for each projection type.}
  \label{fig:analysis-profile}
  \vspace{-10pt}
\end{figure*}
\noindent\textbf{(2) Structured allocation across the backbone.}
Fig.~\ref{fig:analysis-profile} further shows that the generated updates are
not distributed uniformly throughout the frozen generator. In the
single-stream blocks, \texttt{to\_qkv\_mlp\_proj} receives larger updates in
early and middle layers and gradually decreases toward later layers, whereas
\texttt{to\_out} follows a distinct profile with a sharp increase in the final
block. Double-stream blocks also exhibit different layerwise patterns between
the conditioning and image branches. Across projection types,
\texttt{to\_q} and \texttt{to\_qkv\_mlp\_proj} receive the largest relative
updates, while value-related projections are consistently smaller. These
patterns reveal structured allocation across both network depth and projection
type, suggesting that the parameter generator selectively modulates different
components of the frozen backbone rather than applying uniform adaptation.

\section{Conclusion}
\label{sec:conclusion}
We present \methodname, an instruction-conditioned parameter generation framework that repurposes a pretrained image generator for dense visual perception while keeping both the multimodal understanding model and generative backbone frozen. Given an image--instruction pair, \methodname maps its multimodal representation to transient low-rank updates that specialize the frozen generator for the requested prediction. This design enables different predictor behaviors to be instantiated directly from natural-language instructions without task-specific heads, explicit task identifiers, or stored adapters. Through a unified RGB interface, \methodname supports diverse dense prediction tasks, including segmentation, depth estimation, and surface-normal estimation. Our analyses further show that the generated updates exhibit clear task structure while retaining input-dependent variation, highlighting parameter generation as a flexible mechanism for adapting frozen generative backbones to heterogeneous perception objectives.


\section*{Limitations}

ReVision is still bounded by the capabilities of its frozen multimodal
understanding model and image generator, since errors in either backbone cannot
be corrected through backbone fine-tuning. Instance-level parameter generation
also introduces additional computation compared with using a single fixed
adapter. In addition, all dense predictions are expressed through RGB
generation, which is flexible but may be less efficient than task-specific
prediction heads and remains sensitive to the native generative training
objective. This is particularly relevant for tasks that require precise masks
or geometric predictions. Future work could explore stronger frozen backbones,
more efficient parameter generators, and training objectives better aligned
with dense prediction.

\definecolor{textgray}{HTML}{6E6E73}
\makeatletter
\newcommand\applefootnote[1]{%
  \begingroup
  \renewcommand\thefootnote{}%
  \renewcommand\@makefntext[1]{\noindent##1}%
  \footnote{#1}%
  \addtocounter{footnote}{-1}%
  \endgroup
}
\makeatother

\bibliography{ref}
\bibliographystyle{plainnat}
\appendix

\begin{center}
    \Large{\sc\Huge Appendix\\ \normalsize ReVision: Reprogramming Image Generators for Dense Visual Perception}\\
\end{center}
\vskip 4mm
\startcontents[sections]
\vbox{\sc\Large Table of Contents}
\vspace{5mm}
\hrule height .8pt
\vspace{-2mm}
\printcontents[sections]{}{1}{\setcounter{tocdepth}{2}}
\vspace{4mm}
\hrule height .8pt
\vskip 10mm


\section{Implementation Details}
\label{app:impl}
\paragraph{In-context conditioning.}
The reference image and prediction target are independently encoded by the
frozen VAE, patchified, and normalized using the native latent statistics of
the image generator. Following the multi-latent formulation of
\texttt{FLUX.2-klein-base-9B}, the target and reference tokens are concatenated
into a single sequence and assigned different temporal coordinates, with
$T=0$ for the prediction target and $T=10$ for the reference image. The
remaining spatial coordinates follow their native patch positions. The
flow-matching loss is computed only over the target-token portion. The
parameter generator receives only the multimodal representation of the
reference image and instruction and never observes the noisy target.

\begin{table}[!h]
\centering
\tablestyle{6pt}{1.15}
\begin{tabular}{lcc}
\toprule
& stage 1 & stage 2 \\
\midrule
resolution & $256 \times 256$ & $512 \times 512$ \\
steps & 10\,000 & 20\,000 \\
micro-batch / GPU & 24 & 8 \\
gradient accumulation & 1 & 1 \\
optimizer & \multicolumn{2}{c}{AdamW, $\beta = (0.9, 0.999)$, weight decay $10^{-4}$} \\
peak learning rate & $2\times10^{-5}$ & $2\times10^{-5}$ \\
warmup steps & 500 & 500 \\
schedule & \multicolumn{2}{c}{warmup $\to$ constant} \\
gradient clipping & \multicolumn{2}{c}{1.0} \\
precision & \multicolumn{2}{c}{\texttt{bf16}} \\
EMA & \multicolumn{2}{c}{decay $0.999$, no warmup} \\
timestep density & \multicolumn{2}{c}{$t = \sigma(u),\; u \sim \mathcal{N}(0,1)$} \\
\bottomrule
\end{tabular}
\caption{\textbf{Training configuration.} The stage transition loads \emph{only} the
ReVision weights: global step, optimizer state, learning-rate schedule, EMA
and data-order RNG are all reset, so stage $2$ performs a full warmup to the same
peak learning rate rather than inheriting stage $1$'s step counter.}
\label{tab:trainconfig}
\end{table}

\paragraph{Flow matching and timestep sampling.}
Let $\mathbf{e}^{\mathrm{gt}}$ denote the encoded target and
$\varepsilon\sim\mathcal{N}(0,I)$. We construct
\begin{equation}
\mathbf{x}_t
=
(1-t)\mathbf{e}^{\mathrm{gt}} + t\varepsilon,
\qquad
\mathbf{v}
=
\varepsilon-\mathbf{e}^{\mathrm{gt}}.
\end{equation}
We sample $t$ from a standard logit-normal distribution and apply a fixed
resolution-independent shift of $e^{0.8}\approx2.2255$. The generated parameters
depend only on the image--instruction pair and are computed once per sample,
remaining fixed throughout the denoising trajectory.

\paragraph{Optimization and multi-task training.}
Training proceeds in two stages. We first train at $256^2$ for $10$k steps with $500$ warmup steps on semantic segmentation, instance segmentation, depth estimation, and surface-normal estimation. We then transfer the raw non-EMA parameter-generator weights to the $512^2$ stage and train for another $20$k steps with $500$ warmup steps, additionally introducing referring segmentation. The optimizer and learning-rate schedule are restarted between stages. We use AdamW with a peak learning rate of $2.5\times10^{-5}$ and a warmup-stable schedule without terminal decay. At each optimization step, we draw an equal number of samples from every task-specific dataloader and concatenate them into a mixed-task batch. Thus, all tasks contribute equally to each update and are processed jointly in a single batched forward pass through the frozen backbones. We maintain an EMA of the parameter-generator weights with decay $0.999$ and use the EMA checkpoint for evaluation. Detailed hyperparameters are summarized in Tab.~\ref{tab:trainconfig}.
\paragraph{Zero initialization.}
To stabilize optimization, we use standard initialization for all modules except the output projections that generate the LoRA $B$ matrices, which are initialized to zero. As a result, $B_s=0$ and therefore $\Delta W_s=\frac{\alpha}{r}B_sA_s=0$ at every injection site at initialization, regardless of the initial values of $A_s$. The adapted generator is thus functionally identical to the pretrained generator at the start of training. This avoids introducing arbitrary parameter perturbations before the conditioning pathway has learned a meaningful mapping from multimodal representations to parameter updates. During optimization, the parameter generator gradually departs from this zero-update state and learns input-dependent low-rank modifications while the generator remains fixed.

\section{Unified RGB Output and Data Generation}
\label{app:data}

\paragraph{Unified data interface.}
We cast all perception tasks into the same image-to-image training format. Each training sample consists of an RGB input image, a natural-language instruction, and an RGB target image. Task-specific annotations are converted into RGB representations before training, with reserved colors used for invalid or background pixels when applicable. Input images are resized with standard image interpolation, whereas discrete annotations are always resized with nearest-neighbor interpolation to avoid introducing spurious class or instance labels. Samples are resized when necessary and cropped into fixed-resolution square regions without padding. This standardizes spatial dimensions across tasks and ensures that both the input and target satisfy the image generator's resolution constraint that height and width be multiples of $16$. The resulting unified interface allows semantic segmentation, instance segmentation, referring segmentation, depth estimation, and surface-normal estimation to share the same generative training objective.
 
\paragraph{Semantic segmentation.} Semantic annotations are first converted into class-index maps and then rendered as RGB targets using a sample-specific color palette. For each crop, foreground classes are assigned distinct colors sampled without replacement, while background and ignored regions are mapped to black. If more than 32 classes occur in a sample, we retain the classes with the largest pixel areas and map the remaining regions to background. The resulting class-to-color mapping is optionally included in the instruction. Because the palette is redrawn independently for each sample, the model cannot associate a semantic class with a fixed dataset-level color. We further randomize how colors are expressed in language,  RGB, and hexadecimal codes. A crop is discarded if it contains only ignored pixels and therefore provides no valid semantic supervision.

\paragraph{Instance segmentation.} For instance segmentation, we use the original annotations primarily to identify the semantic category and the expected object support, while regenerating high-quality instance masks with SAM~3~\citep{carion2026sam} using the category name as the text prompt. This avoids directly supervising the model with coarse or noisy source masks. Each training row corresponds to one image-category pair, and different instances of the queried category are rendered with different colors. We apply several stages of quality control before constructing the RGB target. First, SAM~3 predictions must satisfy a minimum confidence of $0.55$, and masks with more than $25\%$ of their pixels in the uncertain probability range $(0.3,0.7)$ are rejected. We then remove masks with extreme area ratios or aspect ratios, low mask-to-box occupancy, strong overlap with higher-confidence predictions, or excessive fragmentation. Specifically, valid masks must occupy between $0.0015$ and $0.92$ of the image, have a box aspect ratio below $12$, and fill at least $20\%$ of their bounding box. Mask-level non-maximum suppression is applied at an IoU threshold of $0.7$, and highly fragmented predictions are removed. We additionally compare the union of the regenerated masks with the original annotation. Image-category pairs are rejected when their union IoU or recall falls below dataset-specific thresholds, which prevents incorrect SAM~3 detections from entering the training set. Crops are then constructed around the surviving objects. We include a context crop centered on the union of all instances and up to three object-centric crops around the largest instances. Highly redundant crops with box IoU above $0.85$ are removed. Instances for which less than $25\%$ of the original mask remains inside a crop are treated as background rather than as valid truncated objects. Each surviving instance is assigned a distinct color, with colors selected to be well separated in perceptual RGB space. When the instruction explicitly specifies the background color, it is included in the same separation constraint. After resizing the crop to the final training resolution, we perform a final consistency check and reject samples in which small instances disappear or the number of visible instance colors no longer agrees with the instruction. This ensures that the textual description and rendered supervision remain synchronized. 

\paragraph{Referring segmentation.} Referring segmentation uses the same SAM~3~\citep{carion2026sam}-based mask regeneration and object-centric cropping pipeline, but the segmentation prompt is the referring expression itself rather than a category name. We first sanitize the text before running SAM~3. Truncated expressions containing ellipses, dangling function words, purely deictic phrases, excessively long descriptions, and duplicated expressions are removed. We further require each cleaned expression to correspond to a unique object identity. After SAM~3 filtering, an expression is retained only if exactly one valid instance remains. This provides an additional visual uniqueness check beyond the source annotation. Crops must also preserve a sufficient fraction of the referred object, since a strongly truncated target is no longer a faithful visual grounding of the original expression. The target is represented as a two-color RGB mask, with the referred object assigned a foreground color and all remaining pixels assigned a background color. Both colors may be specified in the instruction and are used directly to render the target. Images appearing in the SA-Co/Gold evaluation set are excluded from this training source to prevent leakage into the zero-shot evaluation. 

\paragraph{Relative depth estimation.} We encode relative depth as an RGB image using a continuous one-dimensional color code $\Gamma$ defined along a Hamiltonian path through the eight vertices of the RGB cube. For a normalized scalar $\rho\in[0,1]$, we use \begin{equation} \Gamma(\rho) = v_j + (7\rho-j)(v_{j+1}-v_j), \qquad j=\lfloor 7\rho\rfloor , \end{equation} where adjacent vertices differ in only one RGB channel. This construction gives a continuous and invertible mapping between a scalar relative depth value and an RGB target. Normalization is performed independently for each crop rather than for the entire source image. Let $\ell$ and $h$ denote the $2$nd and $98$th percentiles of valid depth values within the crop. We compute \begin{equation} u = \operatorname{clip} \left( \frac{x-\ell}{h-\ell}, 0, 1 \right), \qquad \rho=\tau+(1-\tau)u , \end{equation} where the orientation of $x$ is chosen such that $u$ increases with distance. We use $\tau=0.005$ to reserve pure black $(0,0,0)$ exclusively for invalid pixels. The per-image normalization is important because the model observes only the cropped image during training; using statistics from regions outside the crop would introduce an unobservable dependence into the target. We remove crops with insufficient geometric supervision. A crop is discarded when valid depth occupies less than $20\%$ of its pixels. We also remove crops whose robust relative depth variation satisfies \begin{equation} \frac{h-\ell}{\max(|h|,|\ell|)} < 0.01 . \end{equation} This prevents nearly planar or constant-depth regions from being stretched across the full color range, which would amplify small sensor or reconstruction noise into artificial supervision. Since this representation intentionally removes absolute scale and shift, predictions are aligned to ground truth with a least-squares scale-and-shift transformation during evaluation. 

\paragraph{Surface-normal estimation.} Surface normals are represented directly as RGB images by mapping the three components of a unit normal from $[-1,1]$ to $[0,255]$. Predicted RGB values are mapped back to $[-1,1]$ and renormalized to unit length, while invalid pixels are represented by black. Normals from different sources are converted to a common camera coordinate convention with $X_{\mathrm{cam}}$ pointing right, $Y_{\mathrm{cam}}$ pointing down, and $Z_{\mathrm{cam}}$ pointing into the scene. For sources such as TartanAir~\citep{wang2020tartanair} that provide depth but not surface-normal labels, we derive normals directly from depth. Each pixel is back-projected into camera-space 3D using the camera intrinsics, and a local plane is fitted within a $5\times5$ neighborhood. Neighbors whose relative depth differs from the center pixel by more than $3\%$ are excluded to avoid fitting across occlusion boundaries. The normal is obtained from the eigenvector associated with the smallest eigenvalue of the local covariance matrix and is oriented consistently toward the camera. Pixels with invalid depth or insufficient valid neighbors are marked as invalid. Unlike relative depth, normal targets do not require per-crop scale normalization, and valid crops are retained without an additional global quality threshold. 

\paragraph{Instruction construction.} Instructions are sampled from hand-written task-specific templates. For geometry tasks, templates provide different natural-language formulations of the same prediction objective, with the depth instructions explicitly using relative or normalized terminology rather than metric-depth language. For segmentation tasks, instructions can additionally contain the class-to-color, instance-color, or foreground-background mapping used to construct the target. Semantic segmentation includes the color mapping with probability $0.8$, while a fraction of the remaining samples contain only the task instruction.  Most importantly, the colors stated in an instruction are always the exact colors used to render the corresponding target pixels. This prompt-target consistency is enforced after all cropping and resizing operations and forms a final quality-control criterion for the segmentation datasets.

\begin{figure*}[!t]
    \centering
    \includegraphics[width=\linewidth]{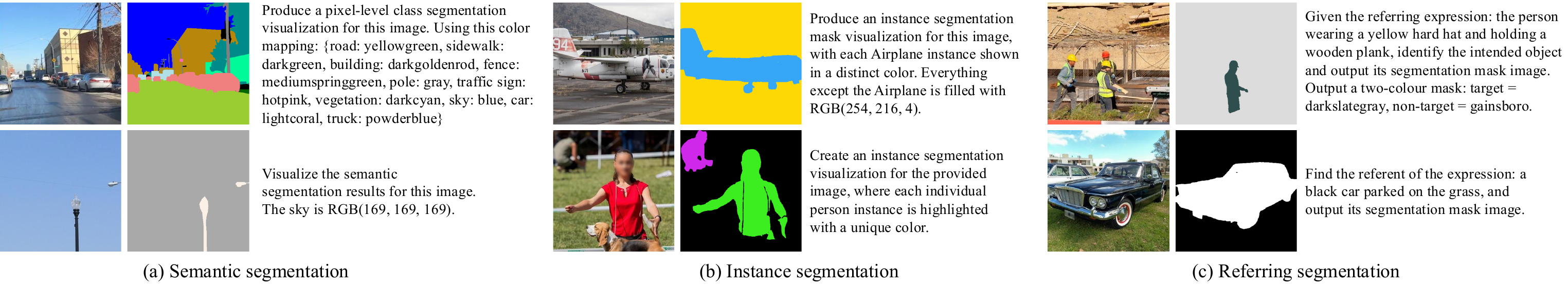}
\caption{\textbf{Examples from the datasets used for different segmentation tasks.}}
    \label{fig:e1}
\end{figure*}

\begin{figure*}[!t]
    \centering
    \includegraphics[width=0.8\linewidth]{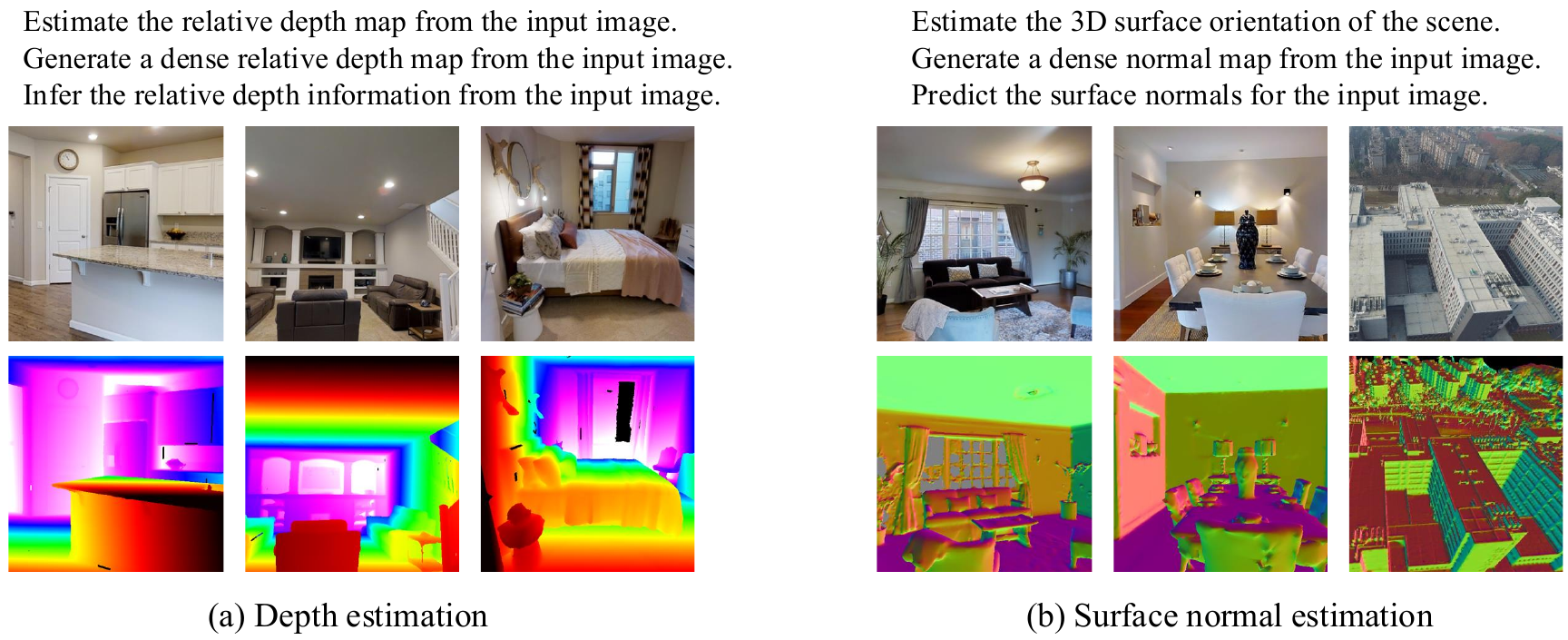}
\caption{\textbf{Examples from the depth and surface-normal datasets.}}
    \label{fig:e2}
\end{figure*}

The data sources for all tasks are summarized in Tab.~\ref{tab:corpus}, with representative examples visualized in Fig.~\ref{fig:e1} and Fig.~\ref{fig:e2}.
\begin{table}[!t]
\centering
\small
\caption{\textbf{Training data sources.}
Datasets and sample counts used in the $256^2$ and $512^2$ training stages.}
\setlength{\tabcolsep}{3pt}
\renewcommand{\arraystretch}{1.08}
\begin{tabularx}{\columnwidth}{@{}p{0.21\columnwidth}Xcc@{}}
\toprule
Task & Training sources & $256^2$ & $512^2$ \\
\midrule
Semantic seg.
& ADE20K, COCO-Stuff-164K, Mapillary Vistas v2, BDD100K
& 0.69M & 1.0M \\

Instance seg.
& LVIS, Open Images V7, GranD
& 2.28M & 7.6M \\

Referring seg.
& GranD
& 0 & 3.0M \\

Relative depth
& Omnidata (Hypersim, BlendMVS, Replica, Replica-GSO, Taskonomy), TartanAir
& 1.7M & 2.7M \\

Surface normals
& Omnidata (Hypersim, BlendMVS, Replica, Replica-GSO, Taskonomy), TartanAir
& 1.8M & 2.9M \\
\midrule
Total & & 6.47M & 17.2M \\
\bottomrule
\end{tabularx}
\label{tab:corpus}
\end{table}


\section{Model Details}
\label{app:impl2}

\paragraph{LoRA injection sites.}
We use \texttt{FLUX.2-klein-base-9B}~\citep{flux2024} as the frozen base generator. It contains 8 double-stream blocks followed by 24 single-stream blocks. The parameter generator predicts LoRA parameters for the attention-related linear projections in the backbone. Specifically, each double-stream block contains eight injection sites:
\texttt{to\_q}, \texttt{to\_k}, \texttt{to\_v}, and
\texttt{to\_out.0} for the image stream, and \texttt{add\_q\_proj}, \texttt{add\_k\_proj}, \texttt{add\_v\_proj}, and \texttt{to\_add\_out} for the conditioning stream. Each single-stream block contains two injection sites, \texttt{to\_qkv\_mlp\_proj} and \texttt{to\_out}. This gives a total of $8 \times 8 + 24 \times 2 = 112$ LoRA injection sites. For each site, the parameter generator produces the corresponding LoRA-$A$ and LoRA-$B$ matrices, while the base-model parameters remain frozen.

\paragraph{ReVision parameter budget.}
The parameter generator consists of an instruction bridge, a learnable query
bank, $N=10$ Transformer blocks, and site-specific decode heads that generate
the LoRA factors for the frozen image generator. We use hidden width $d=1536$,
$24$ attention heads, bottleneck width $m=256$, and LoRA rank $r=16$. For
$S=112$ injection sites, the query bank contains $2Sr=3584$ learnable tokens. A detailed breakdown of the parameter budget is provided in Tab.~\ref{tab:budget}.

\begin{table}[h]
\centering
\caption{\textbf{Parameter budget of the ReVision parameter generator.}
Parameter counts omit linear-layer biases. The per-site decode heads dominate
the budget because they produce LoRA factors at the native widths of the
generator backbone.}
\tablestyle{8pt}{1.15}
\begin{tabular}{lll}
\toprule
Module & Shape & Params \\
\midrule
Instruction bridge
& $\mathrm{Linear}(4096 \to 1536)$
& 6.3M \\
Learnable queries $Q$
& $3584 \times 1536$
& 5.5M \\
Transformer blocks
& $10 \times [\mathrm{Linear}(d \to 9d) + \mathrm{Linear}(4d \to d)]$
& 306.7M \\
Per-site decode heads
& $112 \times [2\,\mathrm{Linear}(d \to m) + \mathrm{LoRA\ decoding}]$
& 599.8M \\
\midrule
Total & & \textbf{918.3M} \\
\bottomrule
\end{tabular}
\label{tab:budget}
\end{table}

\section{Evaluation Protocols}
\label{app:eval}
All evaluations use $30$ sampling steps, a guidance scale of $4.0$, and a fixed random seed of $42$.

\noindent \textbf{Segmentation.} We formulate these tasks as
RGB mask generation and decode predictions by nearest-colour assignment over
the palette specified in the instruction. For semantic segmentation, we
evaluate all $500$ Cityscapes \texttt{val} images at
$1024\times2048$. The instruction specifies the semantic classes present
in the image and their corresponding colours, and predictions are evaluated
against the 19-class ground truth with \texttt{ignore\_index}$=255$. For
instance segmentation, the instruction specifies the target object category,
and the model predicts the corresponding instance mask with all other regions
treated as background. For referring segmentation, the target is specified by
a free-form referring expression, and the model predicts the mask of the
referred region.

\noindent \textbf{Depth.} We evaluate affine-invariant monocular depth following the Marigold protocol. For each image, we generate a single RGB prediction and decode it into relative depth, without ensembling. We then fit a least-squares scale and shift in depth space to align the prediction with the ground truth, and clip the aligned depth to the valid range of each dataset. Evaluation uses the same resolutions, masks, and depth ranges as Marigold, with all metrics computed at the native ground-truth resolution.

\noindent \textbf{Normals.} We evaluate surface-normal estimation zero-shot on NYUv2, DIODE-indoor, ScanNet, and Virtual KITTI 2, without dataset-specific fine-tuning. The predicted RGB output is decoded into unit surface normals, transformed to the evaluation coordinate convention, and resized to the native ground-truth resolution before evaluation.

\section{Analysis Details}
\label{app:analysis}

This section provides the definitions of the statistics used in \secref{sec:analysis}. Unless otherwise stated, all analyses are performed on the generated LoRA updates from the same trained parameter generator. \subsection{Parameter-Space Representation} \label{app:analysis:sketch} Directly vectorizing the generated weight updates is prohibitively expensive. For an injection site $s$, the LoRA update is \begin{equation} \Delta W_s = \frac{\alpha}{r}B_sA_s . \end{equation} We therefore represent each update using a fixed factorized random projection: \begin{equation} S_s = R_{1,s}^{\top}\Delta W_s R_{2,s} = \frac{\alpha}{r} \left(R_{1,s}^{\top}B_s\right) \left(A_s R_{2,s}\right), \label{eq:analysis-sketch} \end{equation} where $R_{1,s}$ and $R_{2,s}$ are fixed Gaussian projection matrices. We use $8\times8$ projections at each of the $112$ injection sites and concatenate their vectorized outputs: \begin{equation} z = \operatorname{Concat}_{s=1}^{112} \operatorname{vec}(S_s) \in \mathbb{R}^{7168}. \end{equation} The same projections are shared across all samples, allowing distances and cosine similarities to be compared in a common parameter space. \subsection{Update Magnitude and Effective Rank} \label{app:analysis:rank} The Frobenius norm of each update can be computed directly from its low-rank factors without constructing $\Delta W_s$: \begin{equation} \lVert \Delta W_s\rVert_F^2 = \left(\frac{\alpha}{r}\right)^2 \operatorname{tr} \left[ (A_s A_s^{\top}) (B_s^{\top} B_s) \right]. \label{eq:analysis-fnorm} \end{equation} Since injection sites have different matrix sizes, comparisons across sites use the relative update magnitude \begin{equation} m_s = \frac{\lVert\Delta W_s\rVert_F} {\lVert W_s\rVert_F}. \label{eq:analysis-relative-norm} \end{equation} We additionally measure the effective rank of each generated update. Let $\{\sigma_j\}$ denote its non-zero singular values and $p_j=\sigma_j/\sum_l\sigma_l$. The effective rank is \begin{equation} \operatorname{erank}(\Delta W_s) = \exp\left( -\sum_j p_j\log p_j \right). \label{eq:analysis-erank} \end{equation} It measures how broadly the available rank-$r$ subspace is utilized and is bounded by $r$.

\subsection{Geometry of Generated Updates}
\label{app:analysis:geometry}

Let $z_n$ denote the parameter sketch of sample $n$, and let
$\mathcal{G}$ denote a group of samples, such as one task. Its \textbf{centroid} and
\textbf{mean distance} to the centroid are
\begin{equation}
    \mu_{\mathcal{G}}
    =
    \frac{1}{|\mathcal{G}|}
    \sum_{n\in\mathcal{G}} z_n,
    \qquad
    d_{\mathcal{G}}
    =
    \frac{1}{|\mathcal{G}|}
    \sum_{n\in\mathcal{G}}
    \lVert z_n-\mu_{\mathcal{G}}\rVert_2 .
    \label{eq:analysis-centroid}
\end{equation}
We report the normalized \textbf{dispersion}
\begin{equation}
    D_{\mathcal{G}}
    =
    \frac{d_{\mathcal{G}}}
         {\lVert\mu_{\mathcal{G}}\rVert_2},
    \label{eq:analysis-dispersion}
\end{equation}
and the \textbf{task-centroid offset} relative to its within-task spread:
\begin{equation}
    O_{\mathcal{G}}
    =
    \frac{
    \lVert\mu_{\mathcal{G}}-\mu_{\mathrm{all}}\rVert_2
    }{
    d_{\mathcal{G}}
    }.
    \label{eq:analysis-offset}
\end{equation}

To characterize the dimensionality of the variation, we compute the
\textbf{participation ratio}:
\begin{equation}
    \operatorname{PR}(\mathcal{G})
    =
    \frac{
    \left(\sum_j\lambda_j\right)^2
    }{
    \sum_j\lambda_j^2
    },
    \label{eq:analysis-pr}
\end{equation}
where $\{\lambda_j\}$ are the eigenvalues of the covariance matrix of the
centered sketches in $\mathcal{G}$. A smaller value indicates that the
variation is concentrated along fewer dominant directions.

\subsection{Gradient Conflict}
\label{app:analysis:gradient}

For each training sample $n$, we compute the gradient of the flow-matching
objective with respect to its generated LoRA parameters and represent it as
$g_n$. For two samples $a$ and $b$, their gradient similarity is measured by
\begin{equation}
    C(a,b)
    =
    \frac{
    \langle g_a,g_b\rangle
    }{
    \lVert g_a\rVert_2
    \lVert g_b\rVert_2
    }.
    \label{eq:analysis-grad-cos}
\end{equation}
For a set of sample pairs $\mathcal{P}$, we report the mean cosine similarity
and the conflict ratio
\begin{equation}
    \operatorname{Conflict}(\mathcal{P})
    =
    \frac{1}{|\mathcal{P}|}
    \sum_{(a,b)\in\mathcal{P}}
    \mathbf{1}\!\left[C(a,b)<0\right].
    \label{eq:analysis-conflict}
\end{equation}
We compute these statistics separately for within-task and cross-task pairs.

\subsection{Parameter-Space Separability}
\label{app:analysis:silhouette}

We use the silhouette score to measure the separation of generated updates by
task identity or image-semantic clusters. For sample $m$, let $a_m$ denote its
mean distance to samples in the same cluster and ${dis}_m$ the minimum mean distance
to samples in any other cluster. We define
\begin{equation}
\mathrm{silh}_m
=
\frac{{dis}_n-a_m}
{\max(a_m,{dis}_m)},
\qquad
\text{silhouette score} = \frac{1}{N}\sum_{n=1}^{N}\mathrm{silh}_m .
\label{eq:analysis-silhouette}
\end{equation}
All distances are computed in the parameter-sketch space, while t-SNE is used
only for visualization.

\section{Visualization Results}
\label{app:q_result}

\begin{figure*}[!t]
    \centering
    \includegraphics[width=0.8\linewidth]{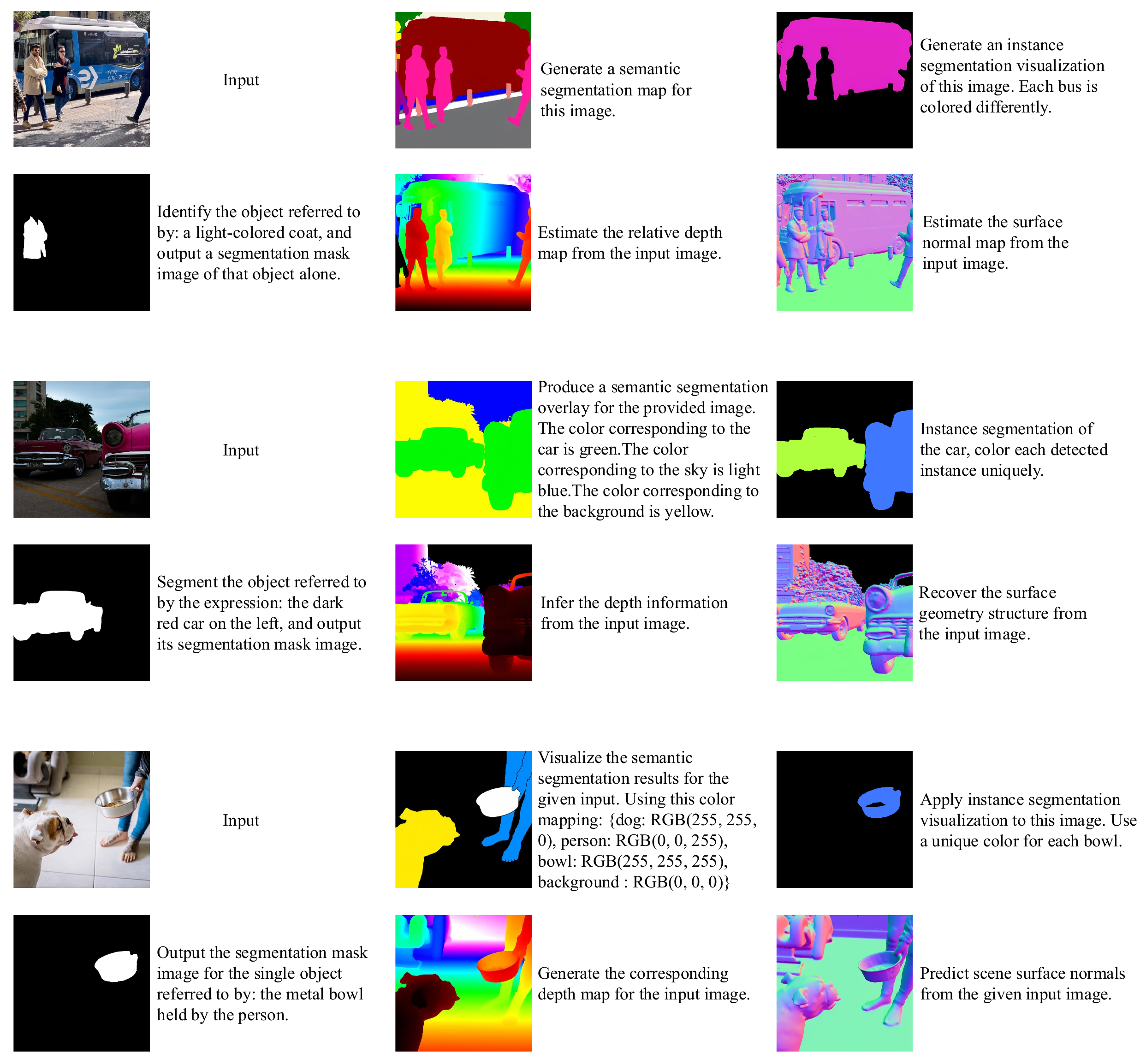}
\caption{\textbf{Qualitative results across dense perception tasks.}
ReVision follows free-form image--instruction inputs to produce pixel-space outputs for semantic segmentation, instance segmentation, referring segmentation, depth estimation, and surface-normal prediction. Different colors and output conventions are specified directly by the instruction.}
    \label{fig:v1}
\end{figure*}
Fig.~\ref{fig:v1} shows representative predictions across five dense
perception tasks. Given the same input image, ReVision can instantiate different
prediction behaviors according to the accompanying natural-language instruction.
For semantic segmentation, the generated masks follow the color mapping specified
in the prompt rather than relying on a fixed dataset-level palette. Instance
segmentation separates individual objects with distinct colors, while referring
segmentation localizes only the region described by the expression. For geometric
prediction, the generated depth maps preserve the relative ordering of foreground
and background regions, and the surface-normal predictions exhibit spatially
coherent orientation changes along object and scene boundaries. These examples
also illustrate that a shared RGB output space can support both discrete
segmentation masks and continuous geometric predictions without task-specific

\section{Preserving Generation and Editing Capabilities}
\label{sec:analysis:generation}

A key property of ReVision is that learning perception does not overwrite the
pretrained generator. The image generator remains frozen throughout training,
and the generated LoRA updates are transient and used only for the current
image--instruction pair. When these updates are removed, ReVision exactly
recovers the original generator parameters and can directly use its native image
generation and editing capabilities. This differs from perception methods that fine-tune the generative backbone,
where perceptual objectives and generative behavior must coexist within the
same optimized parameters. In ReVision, generation and editing remain associated
with the original pretrained model, while dense perception is instantiated
through input-dependent parameter updates. As a result, adding perception does
not require replaying generative data or explicitly balancing generation and
perception objectives during training. Because the backbone parameters are never modified, its native generation and
editing functions remain directly accessible by running the generator without
the generated perception updates.

\clearpage
\newpage

\end{document}